\documentclass[letterpaper, 10 pt, conference]{ieeeconf}
\usepackage{times}
\usepackage[pdftex]{graphicx}
\usepackage{subfigure}
\usepackage{amsmath,amssymb,amsopn,amstext,amsfonts}
\usepackage{cancel}
\usepackage[space]{cite}
\usepackage{pdfsync}
\usepackage{balance}
\usepackage{color}
\usepackage{mathtools}
\usepackage{algpseudocode}
\usepackage{algorithm} 

\algnewcommand\algorithmicforeach{\textbf{for each}}
\algdef{S}[FOR]{ForEach}[1]{\algorithmicforeach\ #1\ \algorithmicdo}

\usepackage{bm}

\usepackage{diagbox}
\usepackage{float}
\usepackage{epstopdf}
\usepackage{pifont}
\usepackage{fixltx2e}
\usepackage{mathrsfs}
\usepackage{multirow}
\usepackage{url}
\usepackage{verbatim}
\usepackage[linkcolor=black,citecolor=black,urlcolor=black,colorlinks=true]{hyperref}

\graphicspath{.}
\DeclareGraphicsExtensions{.png,.jpg,.eps,.pdf}
\IEEEoverridecommandlockouts
\title{\LARGE \bf Estimation and Adaption of Indoor Ego Airflow Disturbance with Application to Quadrotor Trajectory Planning}
\author{Luqi Wang, Boyu Zhou, Chuhao Liu and Shaojie Shen
\thanks{This work was supported by HDJI Lab, RGC General Research Fund Project 16213717 and HKUST Institutional Fund. All authors are with the Department of Electronic and Computer Engineering, Hong Kong University of Science and Technology, Hong Kong, China. {\tt\footnotesize $\{$lwangax, bzhouai, cliuci$\}$@connect.ust.hk, eeshaojie@ust.hk}.}%
}
\begin{document}

\maketitle
\thispagestyle{empty}
\pagestyle{empty}

\begin{abstract}
It is ubiquitously accepted that during the autonomous navigation of the quadrotors, one of the most widely adopted unmanned aerial vehicles (UAVs), safety always has the highest priority. However, it is observed that the ego airflow disturbance can be a significant adverse factor during flights, causing potential safety issues, especially in narrow and confined indoor environments. Therefore, we propose a novel method to estimate and adapt indoor ego airflow disturbance of quadrotors, meanwhile applying it to trajectory planning. Firstly, the hover experiments for different quadrotors are conducted against the proximity effects. Then with the collected acceleration variance, the disturbances are modeled for the quadrotors according to the proposed formulation. The disturbance model is also verified under hover conditions in different reconstructed complex environments. Furthermore, the approximation of Hamilton-Jacobi reachability analysis is performed according to the estimated disturbances to facilitate the safe trajectory planning, which consists of kinodynamic path search as well as B-spline trajectory optimization. The whole planning framework is validated on multiple quadrotor platforms in different indoor environments.
\end{abstract}

\section{Introduction}
\label{sec:introduction}

Nowadays, quadrotors, one of the most popular unmanned aerial vehicles (UAVs), have been widely applied in various scenarios\cite{chong2018act}\cite{manyam2017surveillance}\cite{luqi2018collaborative}. On account of their compact size and flexibility, they can be utilized in various confined indoor spaces, where severe safety issues arise.


To guarantee safety, safe trajectories are generated and tracked by controllers to avoid obstacles \cite{zhou2019robust} \cite{fei2017iros} \cite{liu2017planning}.
In the planning phase, the control error bound is a critical factor that should be considered to ensure safety. 
However, even if the state-of-the-art controllers are applied, the error bound is still hard to determine since it depends on the varying external disturbances during flights.
To provide more accurate control error bounds and better facilitate the trajectory planning, a proper estimation of the disturbance is essential.

\begin{figure}[t]
\begin{center}
\subfigure[\label{fig:wall_4_raw} The scenario of two parallel walls with 4m interval.]
{\includegraphics[width=0.43\columnwidth]{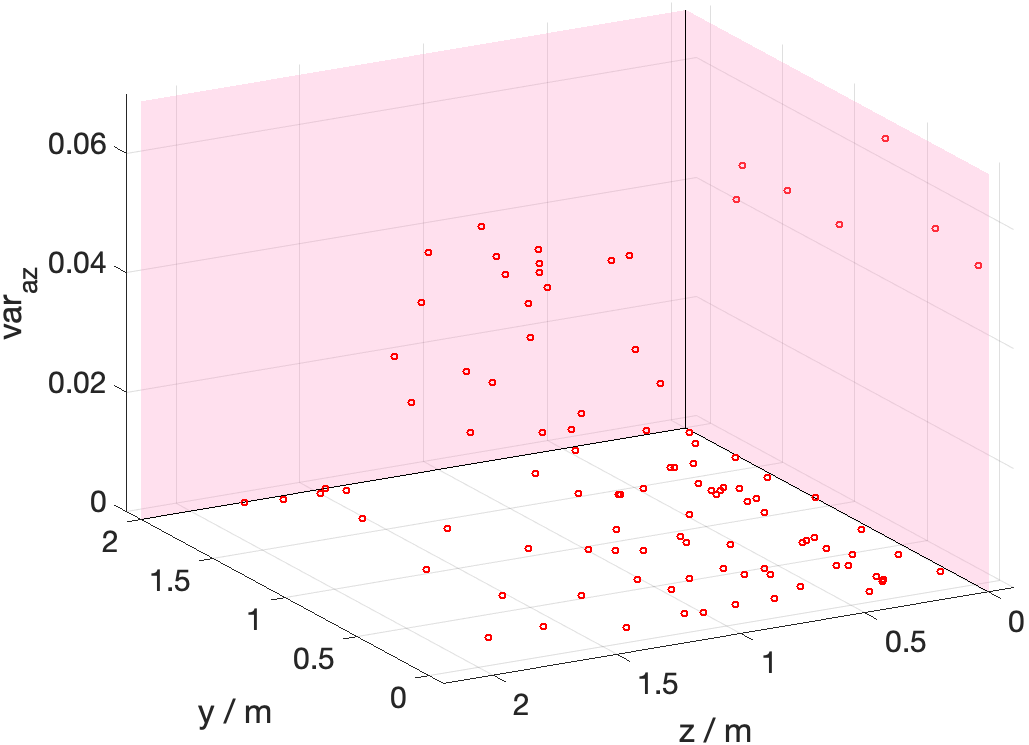}
\includegraphics[width=0.43\columnwidth]{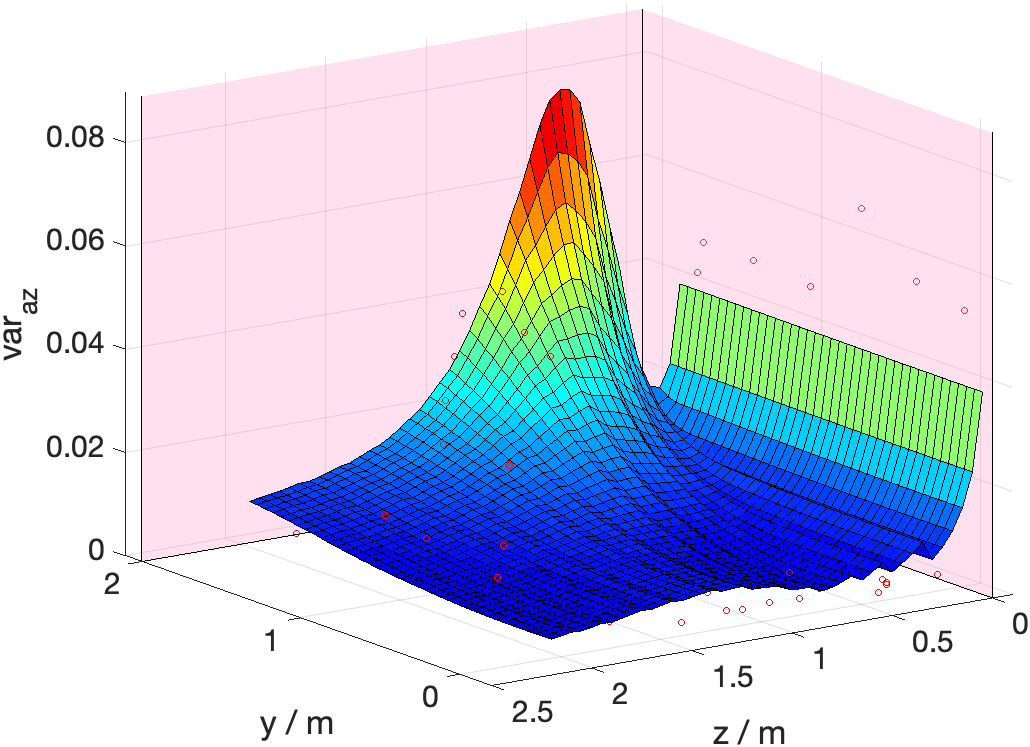}} 
\vspace{-0.3cm}
\subfigure[\label{fig:wall_2_raw} The scenario of two parallel walls with 2m interval.]
{\includegraphics[width=0.43\columnwidth]{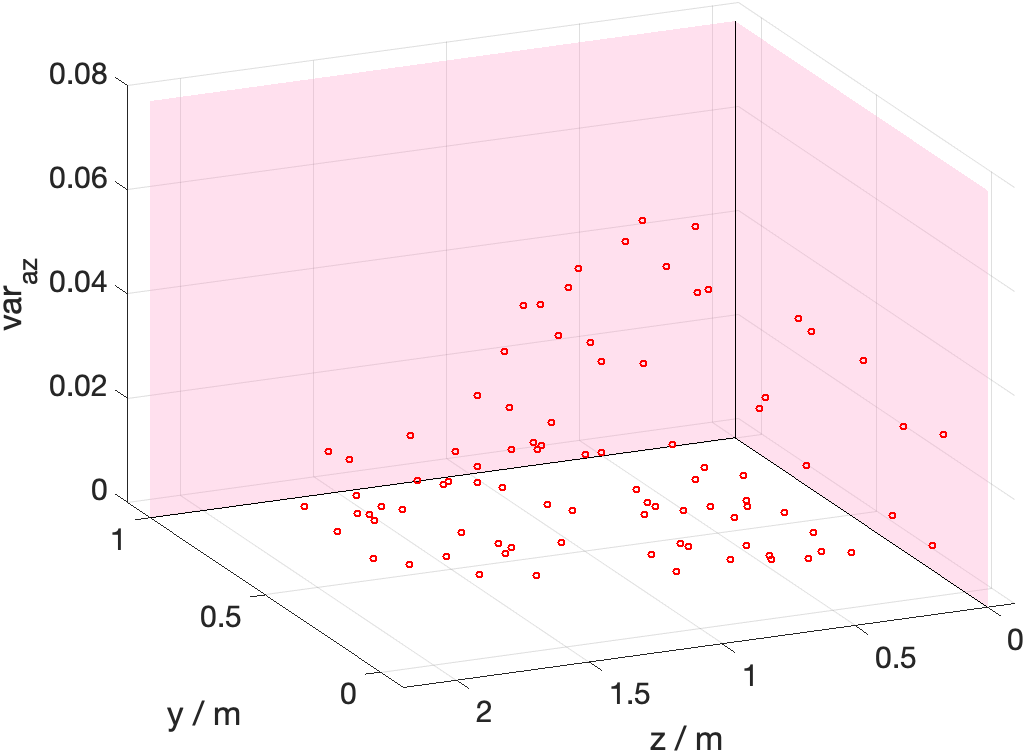}
\includegraphics[width=0.43\columnwidth]{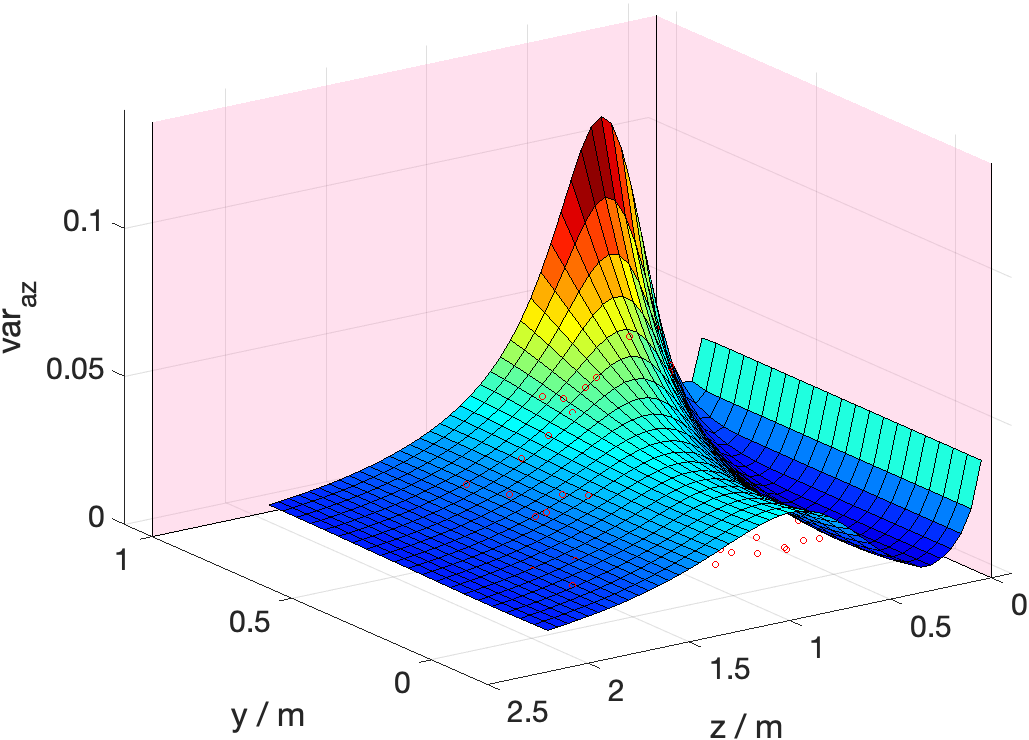}} 
\vspace{-0.2cm}
\subfigure[\label{fig:wall_1_raw} The scenario of two parallel walls with 1m interval.]
{\includegraphics[width=0.43\columnwidth]{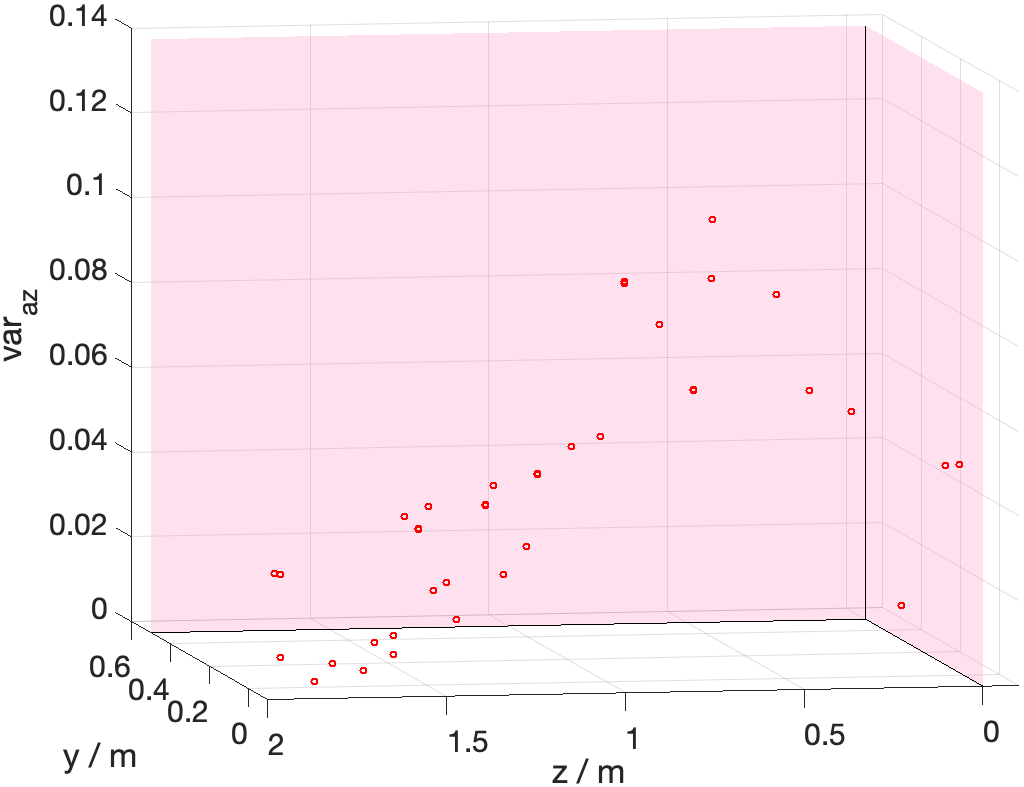}
\includegraphics[width=0.43\columnwidth]{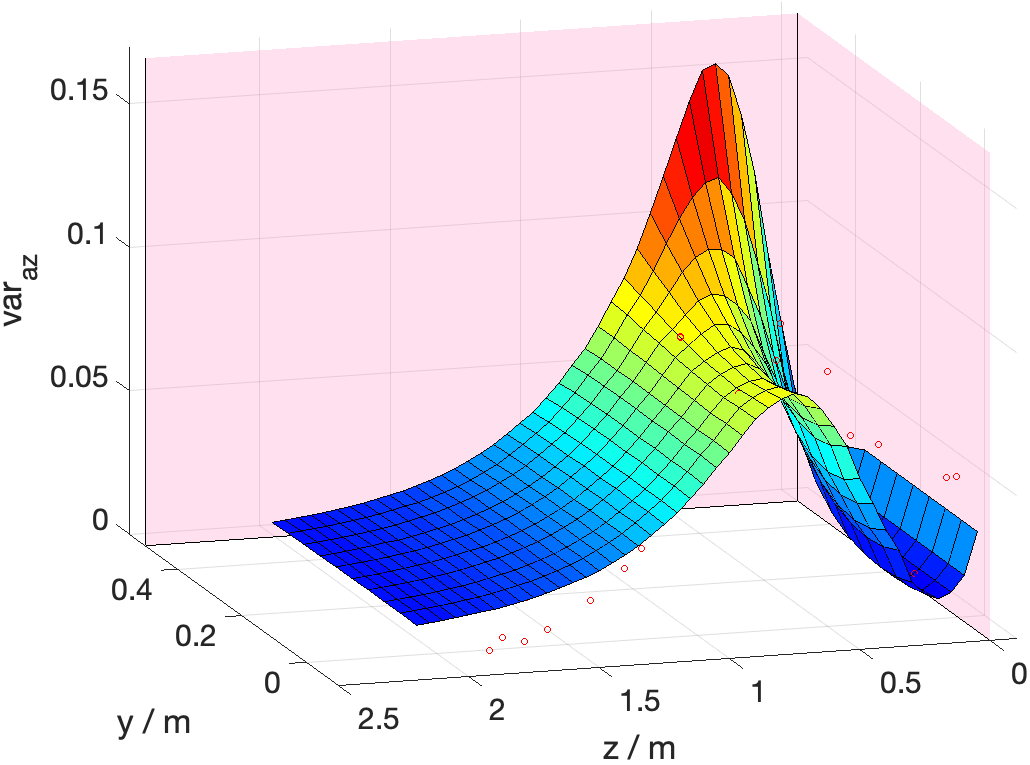}} 
\end{center}
\vspace{-0.3cm}
\caption{\label{fig:wall_raw_bound}The result of the raw acceleration variances and the estimated acceleration variance bounds of the smaller quadrotor shown in Fig. \ref{fig:small_drone_complex} hovering at different position between two parallel walls with different intervals. The experiment setting with is shown in Fig. \ref{fig:data_exp_setup}. The red points are the raw acceleration variances, while the colored surfaces are the estimated acceleration variance upper bound. The pink plane on the right of each of the subplots indicates the ground at 0 height. Meanwhile, the pink plane at the back of each of the subplots indicates the wall at $y = d / 2$. Note that another wall is set at $y = -d / 2$. }
\vspace{-0.6cm}
\end{figure}

Among all the indoor disturbance sources, the most critical source is the airflow from the drone itself. The generated disturbance always follows the quadrotors, meanwhile varies during flights. In particular, the ego airflow from the rotors interacts with obstacles and generates disturbances, causing potential safety issues, especially in confined spaces. During the hover experiments on the quadrotors between the walls with the setup shown in Fig. \ref{fig:data_exp_setup}, the ego airflow disturbance is observed to be significant, especially when the walls have a small interval and the quadrotor hovers near the obstacles. Imprically, the variance of acceleration is closely related to the disturbance force \cite{lee2016robust}, hence adopted as the measurement of the disturbance in this paper. According to the result of the variances shown in Fig. \ref{fig:wall_raw_bound}, the variances near the obstacles can be more than 20 times larger than the variances when hovering far away. Most of the previous planning and control methods, which assume constant control error bounds \cite{zhou2019robust}\cite{hoseong2019frs}, can hardly adapt such disturbance variation. When the constant error bound is underestimated, the significant ego airflow disturbance in cramped areas drives the quadrotor towards the obstacles, which can be detrimental. On the other hand, in the broader areas, the over-estimated error bound can cause a conservative motion planning. 

Therefore, in this paper, to better facilitate safe motion planning, we focus on the estimation and adaption of the indoor ego airflow disturbance caused by the interaction of the rotor flow and the obstacles in the environments, in particular the ground effect and the wall effect. A disturbance field is firstly constructed according to the map information and the proposed model. Then, to ensure safety and robustness, a complete motion planning framework for quadrotors based on kinodynamic path search and B-spline trajectory optimization meanwhile taking account of the control error bounds, which are represented by Hamilton-Jacobi forward reachable sets (FRSs) and computed from the ego airflow disturbance, is proposed. With the proposed method, quadrotors can automatically adapt the ego airflow disturbance so that fly safely and smoothly in complex indoor confined environments.

The contributions of this work are:

\begin{enumerate}
	\item A formulation of the ego airflow disturbance model validated on quadrotors in different environments.
	\item A robust and complete motion planning framework considering the control error bound utilizing the FRSs computed from the estimated disturbances.
	\item A comprehensive integration of the proposed motion planning method, together with state estimation, mapping and control modules into different quadrotor platforms for safe flights.
\end{enumerate}

\section{Related Work}
\label{sec:related_work}
\textbf{Proximity effects between rotor-crafts and obstacles:}
The Proximity effects between rotor-crafts and obstacles have been studied for several decades. One of the most phenomenal effect is the ground effect. The model of the thrust ratio casused by the ground effect is developed by Betz \cite{betz1937ground} and Chesseman \cite{cheeseman1955effect}. The model is widely adopted and analysised by researchers \cite{eberhart2017modeling}\cite{conyers2019empirical} and provides guideline for the applications \cite{powers2013influence}\cite{danjun2015autonomous}. On the other hand, the ceiling effect is proposed by Johnson \cite{johnson2012helicopter} and analyzed by recent researchers \cite{kushleyev2013towards}\cite{conyers2019empirical}. 
Another critical effect is the wall effect. This effect is more complicated and more difficult to formulate. Some simulations \cite{robinson2014computational} and experiments \cite{conyers2019empirical} are conducted to investigate the effect. However, it is extremely hard to come up with a simple formulation as the ground effect.
Although the researches regarding to the proximity have been conducted for such a long time, the main direction is still towards the magnitude of the force or thrust. There is scarce work considering the variance aspect of the proximity effects. In this paper, we mainly focus on the ground effect and wall effect, since as indicated by \cite{kushleyev2013towards}, the magnitude of the ceiling effect is significantly smaller than the ground and wall effect, as well as the airflow is much more stable in the ceiling effect than in the other two effects. The proximity effects are formulated into a disturbance field to provide information for the motion planning framework.

\textbf{Motion planning under disturbances:}
Motion planning for quadrotors under disturbances has also been studied for a long time. One of the most prevalent techniques is using the reachability analysis to compute forward and backward reachable sets \cite{girard2005reachability} \cite{kurzhanski2000ellipsoidal}. The computation of the reachability of non-linear systems with disturbances can be formed into a differential game utilizing the Hamilton\textendash Jacobi\textendash Bellman equation \cite{ian2005hj}, which is also known as HJ reachability analysis. The disturbance and the control input are formulated into a differential game. With a bounded disturbance, the reachable sets of the error states can be computed. The technique is further developed to design robust controller \cite{le2012sequential} and hybrid systems \cite{herbert2017fastrack}\cite{fridovich2018planning}. Another similar method called the funnel libraries is proposed \cite{majumdar2017funnel} to solve more general problems. However, the funnel library requires massive time to pre-compute the funnel to facilitate the real-time planning. Similarly, the HJ reachability analysis also suffers from the curse of dimensionality, which also requires long computation when the dimension increases. The approximation of reachable sets using ellipsoids is a technique to significantly reduce the computation burden \cite{kurzhanski2000ellipsoidal} and can be adopted for motion planning \cite{hoseong2019frs}. 
Nevertheless, most of the previous works only consider the disturbances with a constant bound, which differs from the the real world scenarios. Therefore, in this paper, we decide to adopt the ellipsoidal approximation of the forward reachable sets\cite{kurzhanski2000ellipsoidal} together with the estimated disturbance to facilitate the motion planning framework.
\section{Disturbance field formulation}
\label{sec:disturbance}

\begin{figure}[t]
\begin{center}
{\includegraphics[width=0.3\columnwidth]{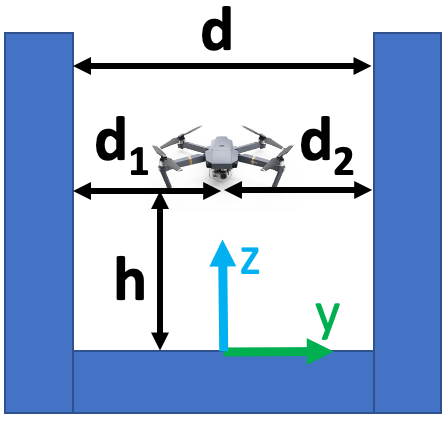}} 
\end{center}
\vspace{-0.4cm}
\caption{\label{fig:data_exp_setup}The parallel wall setup for collecting disturbance data. The quadrotor hovers at a different height $h$ to the ground and different distance $d_1$ and $d_2$ to the two walls. The distance between the two walls varies.}
\vspace{-0.9cm}
\end{figure}

According to the previous researches on the proximity effects \cite{eberhart2017modeling}\cite{conyers2019empirical}, it is shown that the ground effect and the wall effect significantly affect the quadrotors when flying near the obstacles. To address the proximity effects, experiment setups for collecting disturbance data with different wall distances as shown in Fig. \ref{fig:data_exp_setup} are set. It is observed in the experiments that the airflow disturbances are significant during slow flights. Therefore, the disturbance formulation is based on the data during the hover flights which approximates the slow flights.

In consideration of the previous results \cite{eberhart2017modeling}\cite{conyers2019empirical} that the force or acceleration decreases as the quadrotor hovers apart from the ground or the wall, it is empirical that the disturbance variances caused by the turbulent airflow from the rotors have the similar trend, providing the guideline for the disturbance formulation. Meanwhile, since a voxel map is usually adopted for motion planning methods, as well as for the ease of computation, a voxel based acceleration disturbance field model is proposed.

\begin{figure}[t]
\centering
\includegraphics[width=0.28\columnwidth]{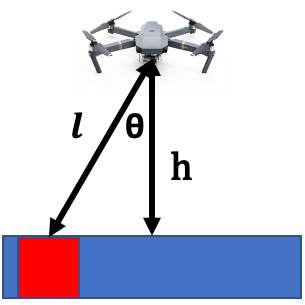}
\vspace{-0.3cm}
\caption{The illustration of the disturbance formulation caused by the ground effect. The red voxel is the voxel being calculated and the total variance of the acceleration is the sum of all the variance contribution of the voxels which can be seen from the quadrotor, within a constant field of view $\theta_g$. \label{fig:ground_form} }
\vspace{-1.4cm}
\end{figure}

As illustrated in Fig. \ref{fig:ground_form}, the contribution of the variance of a voxel caused by the ground effect $\sigma_g$ is formulated as:
\begin{equation}
	\sigma_g=\left\{\begin{array}{ll}
			a^2 \cdot [\frac{\lambda_g}{l^2} + k_{g1} \exp(-k_{g2}l^2)] &  \theta \leq \bar{\theta}_g, l \leq \bar{l}_g \\
			a^2 \cdot [\frac{\lambda_g}{l^2} + k_{g1} \exp(-k_{g2}l^2)] & lsin(\theta) \leq r_0\\
			0 & otherwise
			\end{array}\right.,
\end{equation}
where $a$ is the resolution of the voxel, $l$ is the distance from the voxel to the quadrotor, $h$ is the vertical height of the quadrotor above the voxel. $\theta$ is the angle between the vertical line and the line from the quadrotor to the voxel, $\bar{\theta}_g$ is the preset virtual field of view, $\bar{l}_g$ is the preset length range, $r_0$ is the diagonal radius of the quadrotor; $\lambda_g$, $k_{g1}$ and $k_{g2}$ are the constants to be adjusted according to the experiment data.

\begin{figure}[h]
\begin{center}
\vspace{-0.5cm}
\subfigure[\label{fig:side_form_front} The side view of the quadrotor and a voxel on the wall.]
{\includegraphics[height=0.26\columnwidth]{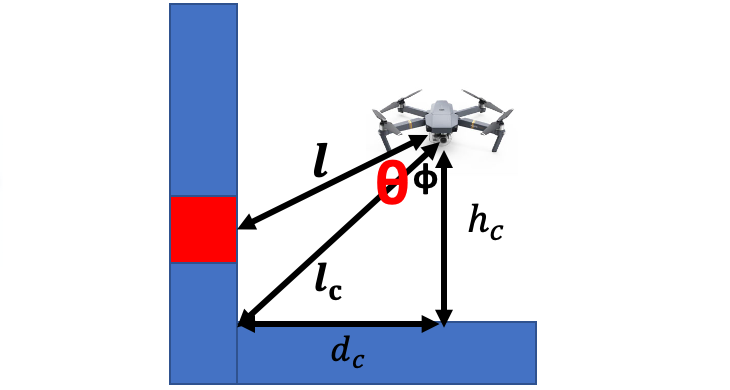}}             
\hspace{0.2cm}
\subfigure[\label{fig:side_form_over} The overhead view of the quadrotor and a voxel on the wall.]
{\includegraphics[height=0.26\columnwidth]{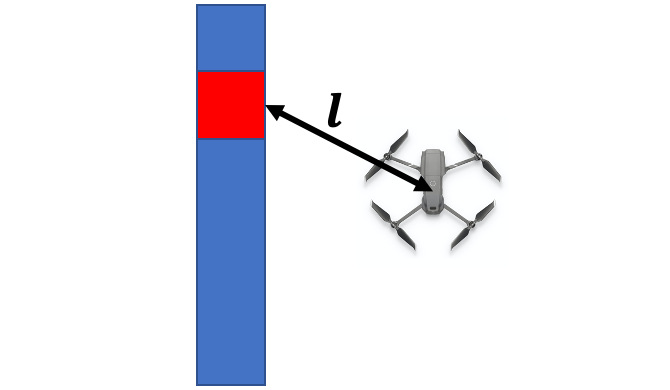}} 
\end{center}
\vspace{-0.4cm}
\caption{\label{fig:side_form}The illustration of the disturbance formulation caused by the wall effect. The red voxel is the voxel being calculated and the total variance of the acceleration is the sum of all the variance contribution of the voxels which can be seen from the quadrotor.}
\vspace{-0.2cm}
\end{figure}

The contribution of the variance of a voxel caused by the wall effect is formulated into two terms: the disturbance caused by the corner $\sigma_c$ and the disturbance caused directly by the wall $\sigma_w$. They are formulated into the following form according to the illustration in Fig. \ref{fig:side_form}:
\begin{equation}
	\sigma_c=\left\{\begin{array}{ll}
			a^2 \cdot (\phi_c-\phi) \frac{\lambda_c}{d_c^2+h^6} &  \phi \leq \bar{\phi}_c, l_c \leq \bar{l}_c \\
			0 & otherwise
			\end{array}\right.,
\end{equation}
\begin{equation}
	\sigma_w=\left\{\begin{array}{ll}
			a^2 \cdot k_{w1} \exp(-k_{w2}l^2) &  \underline{\theta_w} \leq \theta \leq \frac{\pi}{2}, l \leq \bar{l}_w \\
			0 & otherwise
			\end{array}\right.,
\end{equation}
where $\phi$ is the angle between the vertical line and the line from the quadrotor to the corner below the voxel, $l_c$ is the distance from the corner to the quadrotor, $d_c$ is the horizontal distance form the corner to the quadrotor, $h_c$ is the vertical height of the quadrotor above the corner,  $\bar{\phi}_c$ is the preset virtual field of view regarding to the corner, $\bar{l}_c$ is the preset length range regarding to the corner, $\underline{\theta_w}$ is the preset minimal angle regarding to the direct reflection of the wall effect, $\bar{l}_w$ is the preset length range regarding to the wall voxel; $\lambda_c$, $k_{w1}$ and $k_{w2}$ are the constants to be adjusted according to the experiment data.

\section{Control error bound calculation}
\label{sec:hj_reach}

With the estimated disturbance, the HJ-reachability problem is formulated for analyzing the control error bound. The resulting reachable sets of the errors are approximated using ellipsoids for higher computational efficiency \cite{kurzhanski2000ellipsoidal}\cite{hoseong2019frs}, so as to facilitate the safe motion planning mentioned in the following section.

Consider the dynamics of the error state and an affine feedback controller within a small time interval:
\begin{gather}
	\dot{\mathbf{e}}(t)=\mathbf{A}(t)\mathbf{e}(t)+\mathbf{B}(t)(\mathbf{u}(t)-\underline{\mathbf{u}}(t))+\mathbf{D}(t)\mathbf{w}(t) \\
	\mathbf{u}(t) = \mathbf{K}(t)\mathbf{e}(t) + \underline{\mathbf{u}}(t),
\label{eq:ctrl}
\end{gather}
where $\mathbf{e}$ is the error state, consisting of position and velocity errors, $\mathbf{u}$ is the input, $\underline{\mathbf{u}}$ is the nominal input, which is the desired acceleration on the trajectories, $\mathbf{w}$ is the disturbance, and $\mathbf{K}$ is the matrix of the feedback gain.
Denote $\mathbf{\Phi}(t) = \mathbf{A}(t)\mathbf{e}(t)+\mathbf{B}(t)\mathbf{K}(t)$, the FRS becomes:
\begin{equation}
\mathscr{E}(t)=\left\{
\mathbf{y} \left| \begin{array}{l}
\forall \tau \in[0, t], \forall \mathbf{w}(\tau) \in \mathbb{W} \\
\dot{\mathbf{e}}(\tau)=\mathbf{\Phi}(\tau) \mathbf{e}(\tau)+\mathbf{D}(\tau) \mathbf{w}(\tau) \\
\mathbf{e}(0) \in \mathscr{E}(0), \mathbf{y}=\mathbf{e}(t)
\end{array}\right.\right\},
\end{equation}
where $\mathbb{W}$ is the set of the admissible set of the disturbance and $\mathscr{E}_{0}$ is the initial set of the error states.

The solution of the FRS is the following HJ partial differential equation:
\begin{equation}
\begin{gathered}
	\frac{\partial V}{\partial t}+\max_{\mathbf{w} \in \mathbb{W}} (\frac{\partial V}{\partial \mathbf{e}} \cdot \dot{\mathbf{e}}(t))=0 \\
	V(\mathbf{e}, 0)=h(\mathbf{e}(0)),
\end{gathered}
\end{equation}
where $V$ is the value function and $h$ is the convex function satisfies $\mathscr{E}_0 = \{\mathbf{y} | h(\mathbf{y}) \leq 0\}$.

Consider the quadratic initial cost, with positive definite $\mathbf{Q}(0)$
\begin{equation}
	h(\mathbf{e}(0))={\mathbf{e}(0)}^T\mathbf{Q}(0)^{-1}\mathbf{e}(0)-1.
\end{equation}
The initial error state set is equivalent to an ellipsoidal set:
\begin{equation}
	\mathscr{E}(0) = \left\{\mathbf{Q}(0)^{\frac{1}{2}}\mathbf{v} \bigg| ||\mathbf{v}||^2_2 \leq 1 \right\}
\end{equation}

For the disturbance of a channel $i$, the shape matrix of the ellipsoid $\mathbf{Q}_i(t)$ is the solution of a Lyapunov equation:
\begin{equation}
\begin{gathered}
\begin{array}{l}
-\mathbf{\Phi}\left(\mathbf{Q}_{i}(t)-\varepsilon t^{2} I\right)-\left(\mathbf{Q}_{i}(t)-\varepsilon t^{2} I\right) \mathbf{\Phi}^{T} \\
\quad=\exp (-\mathbf{\Phi} t) \mathbf{N}_{i}(t) \exp \left(-\mathbf{\Phi}^{T} t\right)-\mathbf{N}_{i}(t),
\end{array} \\
	\mathbf{N}_{i}(t) = t\bar{w}^2_i\mathbf{D}_i(t)\mathbf{D}_i(t)^T,
\end{gathered}
\end{equation}
where $\varepsilon$ is a positive scalar indicating the conservativeness, $\bar{w}_i$ is the bound of the disturbance of the $i$-th channel, and $\mathbf{D}_i(t)$ is the $i$-th column of the matrix $\mathbf{D}(t)$.

After adopting the conservative Minkowski sum of the ellipsoids of the disturbances of each channel, the shape matrix $\mathbf{Q}(t)$ and be calculated as:
\begin{equation}
\begin{gathered}
\begin{array}{l}
	\mathbf{Q}(t)= \\
	\quad \exp (\mathbf{\Phi} t)\left(\left(1+\frac{b}{a}\right) \mathbf{Q}(0)+\left(1+\frac{a}{b}\right) \mathbf{Q}_{d}(t)\right) \exp \left(\mathbf{\Phi}^{T} t\right),
\end{array} \\
	\mathbf{Q}_{d}(t)=\left(\sum_{i=1}^{n_{w}} \sqrt{tr\left(\mathbf{Q}_{i}(t)\right)}\right)\left(\sum_{i=1}^{n_{w}} \frac{\mathbf{Q}_{i}(t)}{\sqrt{tr\left(\mathbf{Q}_{i}(t)\right)}}\right),
\end{gathered}
\end{equation}  
where $n_w$ is the number of the channels of the disturbance and $a = \sqrt{tr(\mathbf{Q}_0)}$, $b=\sqrt{tr(\mathbf{Q}_d(t))}$.

In this work, since the error bounds represented by the reachable sets are examined in voxel maps, which provides position information about the obstacles, the original ellipsoid consisting of the state of both position and velocity needs to be projected onto the position space. For an ellipsoid of the full error state with shape matrix $\mathbf{Q}$
\begin{equation}
	\mathscr{E}(\mathbf{e}) = \left\{\mathbf{e} \bigg| h(\mathbf{e}) = \mathbf{e}^T\mathbf{Q}^{-1}\mathbf{e}-1 = 0 \right\},
\end{equation}
the projection bound of it on the position space $\mathscr{E}_p$ satisfies the derivative respect to the velocity equals to $\mathbf{0}$ :
\begin{equation}
	\mathscr{E}_p(\mathbf{e}) = \left\{ \mathbf{e} \bigg| \frac{\partial h(\mathbf{e})}{\partial \mathbf{v}} = \mathbf{0} \right\},
\end{equation}
thus eliminating the velocity errors, projecting the original ellipsoids onto the position space for collision checking.
\begin{figure}[t]
\centering
\includegraphics[width=0.7\columnwidth]{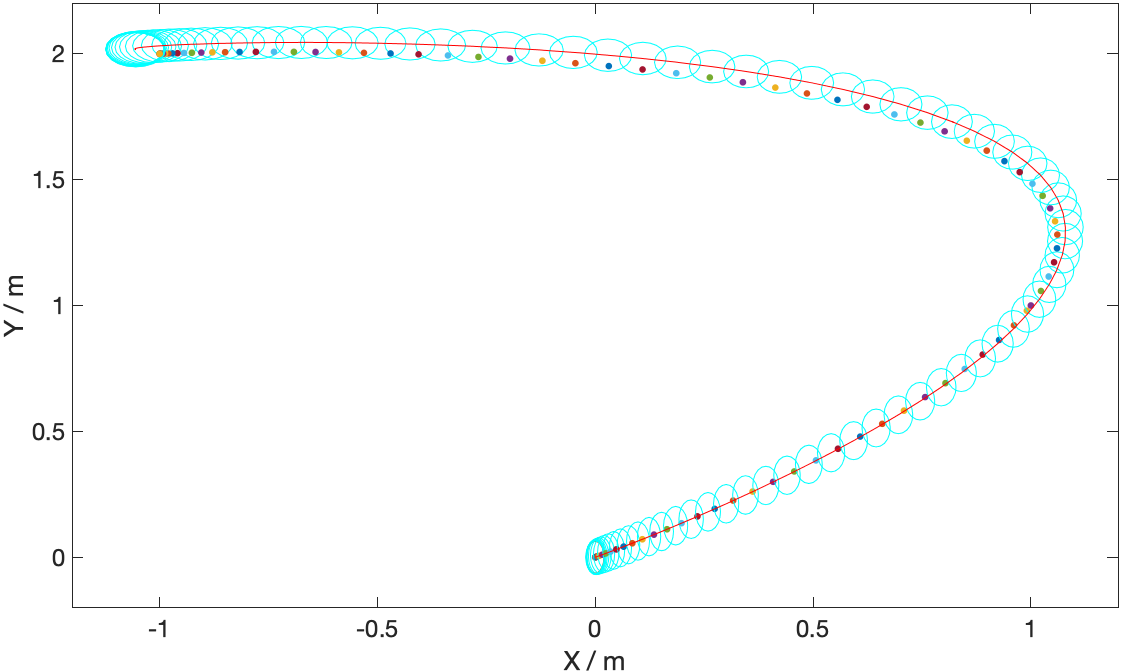}
\vspace{-0.3cm}
\caption{The illustration of the 2-D ellipsoidal reachable sets propagated along a trajectory with constant disturbance bound. The colored dots are the position commands sent out along the trajectory, the red curve is the trajectory of the predicted mean position under the controller, while the blue ellipses are the approximated reachable sets along the propagation. \label{fig:2d_ellipse} }
\vspace{-0.9cm}
\end{figure}

Meanwhile, for the centers of the ellipsoidal error bounds, rather than directly setting them at the positions corresponding to specific time instances on the trajectory, the discrete control commands are taken into account. When a command is processed, the input according to Eq. \ref{eq:ctrl} is computed and executed. Therefore, it is more reasonable to propagate the commands along the trajectory rather than assuming that the centers have zero errors. The illustration of the 2-D ellipsoidal reachable sets propagated along a trajectory with constant disturbances is shown in Fig. \ref{fig:2d_ellipse}. In practice, the FRSs are propagated at the control frequency of 50 Hz and the disturbance in each step is assumed to be constant.

\section{Motion Planning under disturbance}
\label{sec:traj}
To ensure safe flights, a two phase planning method considering the calculated error bound extended from our previous work \cite{zhou2019robust} is proposed.

\subsection{Hybrid A* Search}
\label{sec:hybrid_A*}
For the path searching phase, the hybrid-state A* search \cite{dolgov2010path} with modifications according to the reachability analysis is adopted.
Firstly, the input of the system, which corresponds to the acceleration inputs on each of the axes, is discretized, and multiple primitives are generated according to the state transition equation:
\begin{equation}\label{equ:solution}
	\mathbf{x}(t) = \exp({\mathbf{A}t})\mathbf{x}(0) + 
	\int_{0}^{t} \exp{\mathbf{A}(t-\tau)}\mathbf{B}\mathbf{u}(\tau) \ d\tau.
	\vspace{-0.1cm}
\end{equation}
Multiple time steps are adopted so that multiple layers of primitives are generated. The control cost or the actual cost $\mathcal{G}$ is defined as the combination of the input and the time:
\begin{equation}\label{equ:cost_search}
	\mathcal{G}(T) = \int_{0}^{T} \Vert \mathbf{u}(t) \Vert^{2} dt + \rho T,
\end{equation}
where $\rho$ is a constant that can be adjusted according to the compensation of the time and the input.

The heuristic cost $\mathcal{H}$ is the cost that minimize $\mathcal{G}$ from the searching state to the goal state assuming no obstacles and utilizing the Pontryagin’s minimum principle\cite{mueller2015computationally}:
\begin{equation}
\label{equ:poly}
\begin{gathered}
\begin{bmatrix}
	{\alpha_{\mu}} \\ {\beta_{\mu}}
\end{bmatrix}
= \frac{1}{T^{3}}
\begin{bmatrix}
	 {-12} & {6 T} \\ {6 T} & {-2 T^{2}}
\end{bmatrix}
\begin{bmatrix}
p_{\mu g}-p_{\mu s}-v_{\mu s} T \\ v_{\mu g}-v_{\mu s}
\end{bmatrix} 
\\
\mathcal{H}(T) = \min_{T}(\rho T + \sum_{\mu \in \{x,y,z\}}(\frac{1}{3}\alpha_{\mu}^{2}T^{3} + \alpha_{\mu} \beta_{\mu} T^{2} + \beta_{\mu}^{2} T)),
\end{gathered}
\end{equation}
where $p_{\mu s}$, $p_{\mu g}$ and $ v_{\mu s}$, $v_{\mu g}$are the searching and the goal positions and velocities respectively. 
The total cost used for the hybrid-state A* search $\mathcal{F}$ is simply $\mathcal{G}+\mathcal{H}$.

Before the searching phase, the map is pre-computed for the infinite-time reachable sets, which corresponds to the hover state of the drones, according to the estimated disturbance. The search states are restricted from entering into the unsafe hover positions to prevent possible collisions.

\subsection{B-spline Trajectory Optimization}
\label{sec:b-spline}
Once the hybrid-state A* search completes, the path is parameterized into a B-Spline and the gradient-based optimization starts to generate a smooth and safe trajectory. 

Similar to our previous work \cite{zhou2019robust}, the cost function $f_{total}$ is defined as the weighted sum of the smoothness cost $f_s$, the collision cost $f_c$ and the physical velocity and acceleration constrain cost $f_p$:
\begin{equation}\label{equ:cost}
	f_{total} = \lambda_{s} f_{s} + \lambda_{c} f_{c} + \lambda_{p} f_{p}.
\end{equation}

The smoothness cost $f_s$ is set to be the elastic band cost \cite{quinlan1993elastic} \cite{zhu2015convex}, while the approximation for the third order derivatives of the positions of the control points are adopted, which is closely related to the jerk cost on the trajectory:
\begin{equation}\label{eq:elastic} 
	f_{s} 
	= \sum\limits_{i=p_b}^{N-p_b} \Vert -\mathbf{Q}_{i} + 3\mathbf{Q}_{i+1} - 3\mathbf{Q}_{i+2} + \mathbf{Q}_{i+3} \Vert^{2},
\end{equation}
where $p_b$ is the order of the B-spline and $p_b \geq 3$, $\mathbf{Q}_{i}$ represents the position of the $i$-th control point, $N$ represents the total number of the control points.

The collision cost $f_c$ is evaluated differently from our previous work. Firstly, the disturbance estimated in Sec. \ref{sec:disturbance} as well as the HJ-reachability analysis in Sec. \ref{sec:hj_reach} are adopted to propagate the ellipsoidal error bounds along the trajectory. Then, the half long axes of the ellipsoids $r$ are extracted. The collision cost is defined as the line integral of the squared distance difference value along the trajectory \cite{fei2017iros}:
\begin{equation}\label{equ:colli}
\begin{split}
	f_{c} & = \int_{0}^{T} F_{c}(d(\mathbf{p_p}(t))) \| \mathbf{v_p}(t) \| dt  \\ & =  \sum\limits_{i=0}^{T / \delta t} F_{c}(d(\mathbf{p_p}( \tau_i ))) \| \mathbf{v_p}(\tau_i) \| \delta t,
\end{split}
\end{equation}
where $\delta$ is the time of the propagation steps, $\tau_i = i\delta t$, $\mathbf{p_p}(t)$ and $\mathbf{v_p}(t)$ are the position and the velocity predicted along the trajectory as mentioned in Sec. \ref{sec:hj_reach} at time $t$, meanwhile $ d(\mathbf{p_p}(t)) $ is the distance between $\mathbf{p_p}(t)$ and the closet obstacle. Furthermore, the function $F_c$ is defined as:
\begin{equation}\label{eq:potential}
	F_{c}(\mathbf{p_p}(t)) = \left\{
	\begin{array}{cl}
	(d(\mathbf{p_p}(t)-r(t))^{2} & d(\mathbf{p_p}(t)) \le r(t) \\
	0 & d(\mathbf{p_p}(t)) > r(t)
	\end{array}
	 \right.,
\end{equation}
where $r(t)$ is the safety margin, which is set to be the maximum axis of the ellipsoid at time $t$ from the HJ-reachability analysis. Since the points on the predicted trajectory rather than the control points are adopted for cost computation, the requirement of collision-free convex hull from our previous work \cite{zhou2019robust} on the limit of the distance between the control points is relieved for the trajectory.

Similar to our previous work \cite{zhou2019robust}, according to the convexity of the B-spline, the physical constrain cost $f_p$ is penalized on the control points exceeding the maximum allowable velocity $v_{max}$ and acceleration $a_{max}$, which can also bound the velocity and the acceleration on the trajectory. The cost of a velocity $v_{\mu}$ is defined as:
\begin{equation}\label{key}
	F_{v}(v_{\mu}) = \left\{
	\begin{array}{ccl}
	(v_{\mu}^{2} - v_{max}^{2})^{2} & & v_{\mu}^{2} > v_{max}^{2} \\ 
	0 & & v_{\mu}^{2} \le v_{max}^{2}
	\end{array}
	\right.,
\end{equation}
where $\mu \in \{x,y,z\}$ and the acceleration cost is defined similarly. The total physical cost $f_p$ is defined as:
\begin{equation}\label{equ:kino}
	f_{p} = \sum\limits_{\begin{subarray}{c}
		\mu \in \\ \{x,y,z \}
	\end{subarray}} (\sum\limits_{i=p_b-1}^{N-p_b} F_{v}(\mathbf{V}_{i\mu}) + \sum\limits_{i=p_b-2}^{N-p_b} F_{a}(\mathbf{A}_{i\mu})).
\end{equation}
In practice, the wight on the collision cost $\lambda_{c}$ is chosen to be much larger than the other two weights $\lambda_{s}$ and $\lambda_{p}$.

\begin{figure}[t]
\begin{center}
\subfigure[\label{fig:large_drone_between_walls} The larger quadrotor platform using 7-inch propellers hovering at between two parallel walls with 1m interval for flight data collection.]
{\includegraphics[width=0.46\columnwidth]{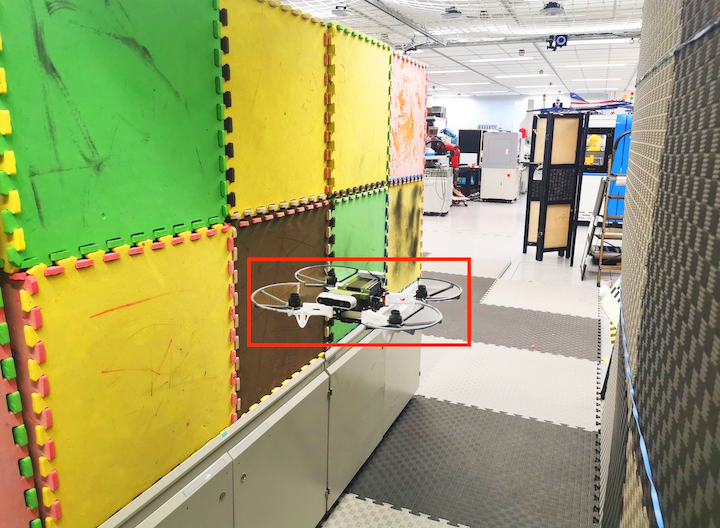}}
\hspace{0.1cm}
\subfigure[\label{fig:small_drone_complex} The smaller quadrotor platform using 5-inch propellers hovering in a complex environment for the verification of the disturbance model.]
{\includegraphics[width=0.46\columnwidth]{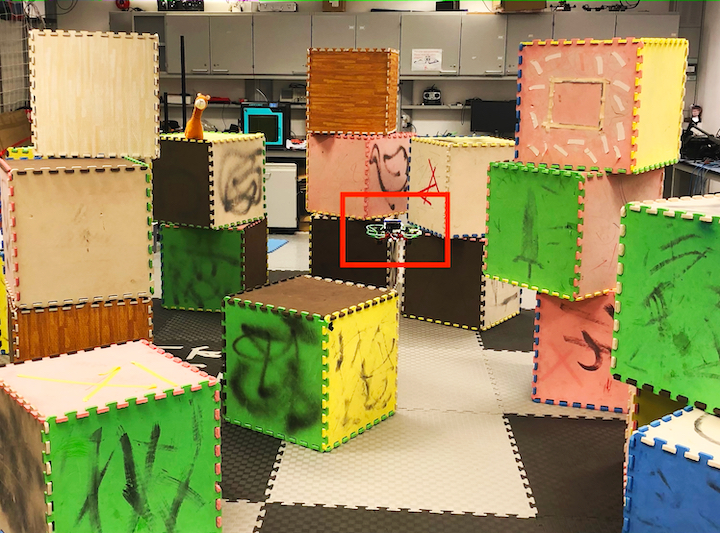}} 
\end{center}
\vspace{-0.5cm}
\caption{\label{fig:drone_hover}The two quadrotor platforms hovering in different environment setting for the flight data collection and the disturbance model validation.}
\vspace{-0.9cm}
\end{figure}

\section{Experiments and Results}
\label{sec:results}

\subsection{Implementation Details}
\label{subsec:implementation}

The proposed disturbance estimator and motion planning method in this paper is implemented in C++11 standard with an open-source non-linear optimization solver NLopt \footnote{\url{https://github.com/stevengj/nlopt}}. The visual-inertial state estimation module VINS-Fusion \cite{qin2019a} and the deformable dense mapping module Dense Surfel Mapping \cite{wang2019real} are adopted. For the experiments in complex environments with various obstacles, the map of the environment is firstly constructed using the Dense Surfel Mapping module as well as the VINS-Fusion module with loop closure detection, preventing the visual-inertial odometry (VIO) from severe drift, which can cause unexpected crash during the subsequent flight. Then, during the flight, the planning algorithm runs fully onboard and the commands are rectified according to the loop closure results to ensure the correct relative position between the obstacles and the quadrotors. In order to validate the proposed disturbance estimation and motion planning method, two quadrotors with different size and configurations as shown in Fig. \ref{fig:drone_hover} are adopted and tested individually in the experiments.

\begin{figure}[t]
\begin{center}
\subfigure[\label{fig:map_1_with_cmd} The constructed map of the complex environment. The black dots indicates the hover positions.]
{\includegraphics[height=0.31\columnwidth]{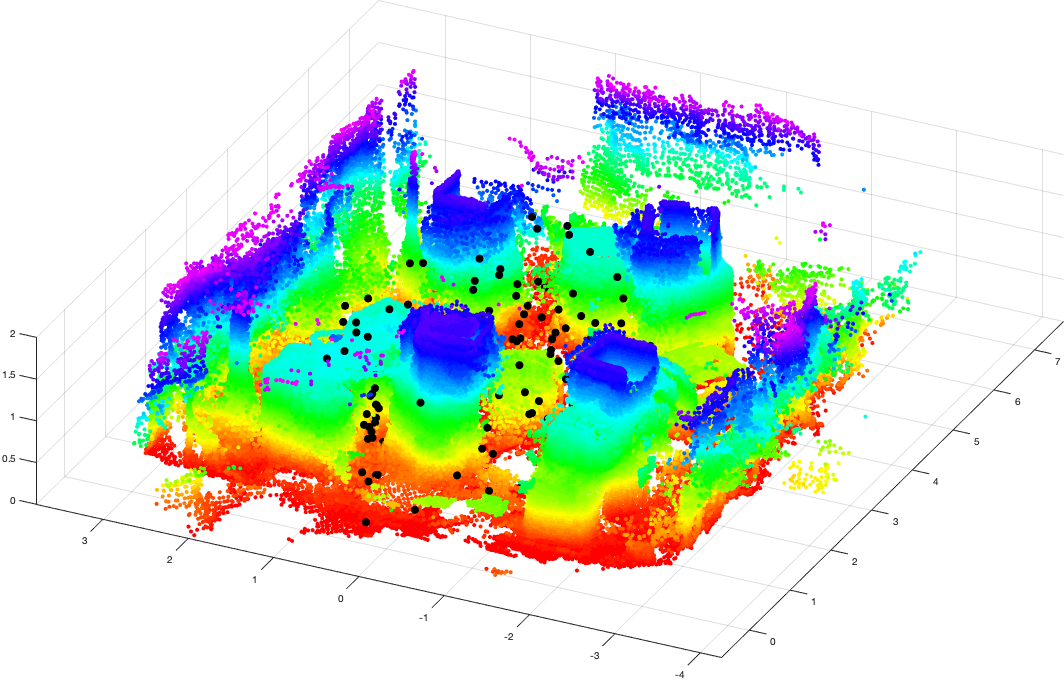}}
\subfigure[\label{fig:map_1_var} The estimated variance bound of the z-axis acceleration at 1.0m height. The color code indicates the magnitude of the variance.]
{\includegraphics[height=0.31\columnwidth]{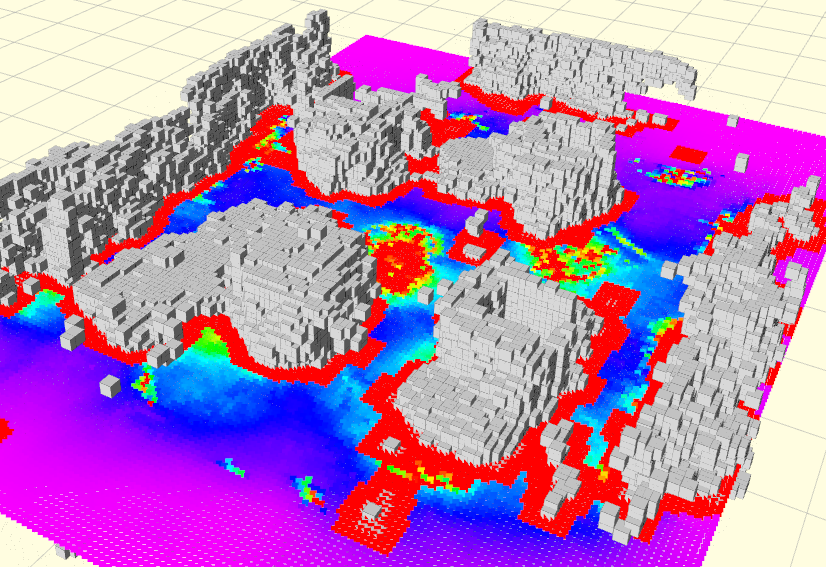}}
\subfigure[\label{fig:large_map_var} The estimated variance bound and the real variance of the z-axis acceleration at the hover positions in Fig. \ref{fig:map_1_with_cmd}.]
{\includegraphics[width=0.76\columnwidth]{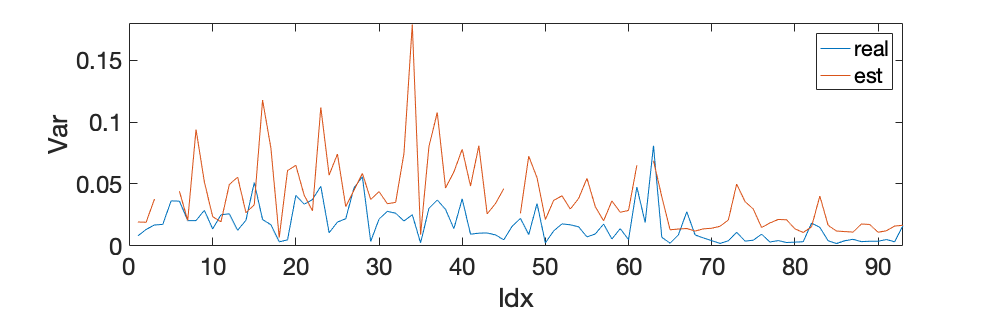}} 
\end{center}
\vspace{-0.5cm}
\caption{\label{fig:hover_result_large}The variance result of the z-axis acceleration of the larger quadrotor in Fig. \ref{fig:large_drone_between_walls} in a complex environment.}
\vspace{-0.4cm}
\end{figure}

\begin{figure}[t]
\begin{center}
\subfigure[\label{fig:map_0_with_cmd} The constructed map of the complex environment. The black dots indicates the hover positions. The color code indicates the height.]
{\includegraphics[height=0.32\columnwidth]{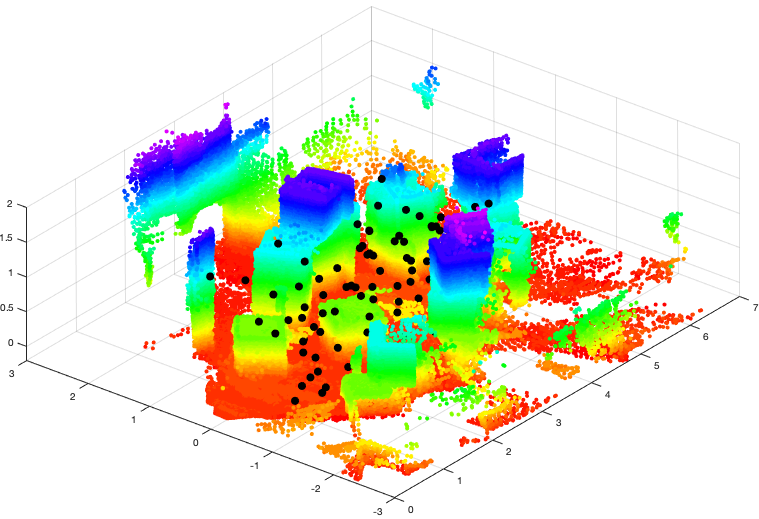}}
\subfigure[\label{fig:map_0_var} The estimated variance bound of the z-axis acceleration at 0.5m height. The color code indicates the magnitude of the variance.]
{\includegraphics[height=0.32\columnwidth]{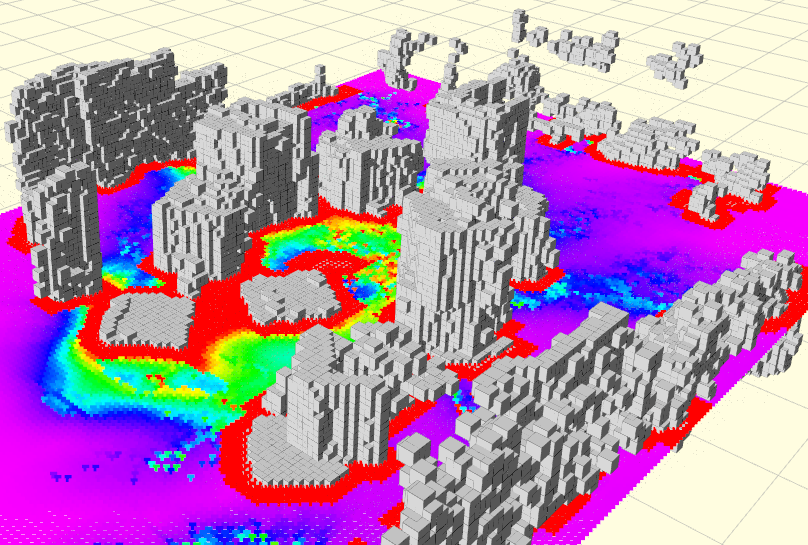}}
\subfigure[\label{fig:map_var} The estimated variance bound and the real variance of the z-axis acceleration at the hover positions in Fig. \ref{fig:map_0_with_cmd}.]
{\includegraphics[width=0.76\columnwidth]{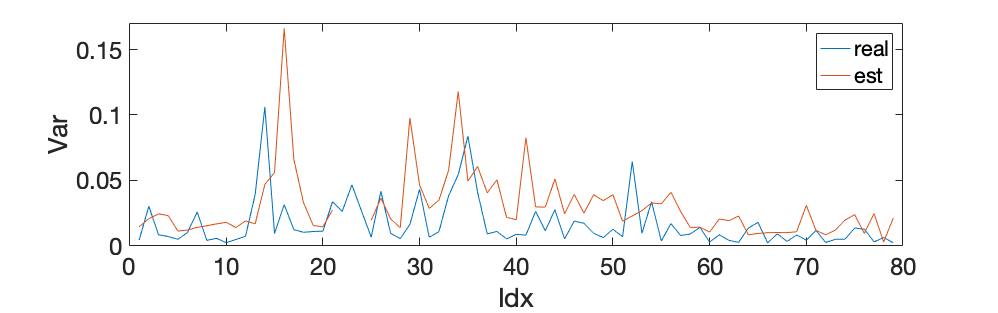}} 
\end{center}
\vspace{-0.5cm}
\caption{\label{fig:hover_result_small}The variance result of the z-axis acceleration of the smaller quadrotor in a complex environment.}
\vspace{-1.3cm}
\end{figure}

\subsection{Disturbance Formulation}
\label{subsec:dist_form_result}
Initially, three sets of long-time hover experiments between two parallel walls with distance of 4m, 2m and 1m, as shown in Fig. \ref{fig:data_exp_setup}, are conducted for both of the quadrotors for flight data collection. After applying the disturbance model mentioned in Sec. \ref{sec:disturbance} on both of the quadrotors, despite of the possible experiment error, at least 85\% of the raw variance data are bounded by the bound surface for both of the quadrotors in each of the setup, which is generated by the same parameters of the proposed disturbance model. Part of the results for the smaller quadrotor are shown in Fig. \ref{fig:wall_raw_bound}.

Moreover, the model is further validated with both of the quadrotors hovering in complex environments. One of the shots during the experiments is shown in Fig. \ref{fig:small_drone_complex}. The two quadrotors hovers in different environments as shown in Fig. \ref{fig:map_1_with_cmd} and \ref{fig:map_0_with_cmd} and obtain the flight data partly shown in Fig. \ref{fig:large_map_var} and \ref{fig:map_var}. In spite of the noisy mapping result, the estimated variance of the acceleration can still bound more than 97\%, 92\%, 92\% and 82\%, 89\%, 82\% of the real fight data in three axes for the two drones, which indicates the effectiveness of the proposed disturbance estimation method.

\begin{figure}[t]
\vspace{-0.1cm}
\begin{center}         
\subfigure[\label{fig:large_composite} Snapshot.]
{\includegraphics[height=0.43\columnwidth]{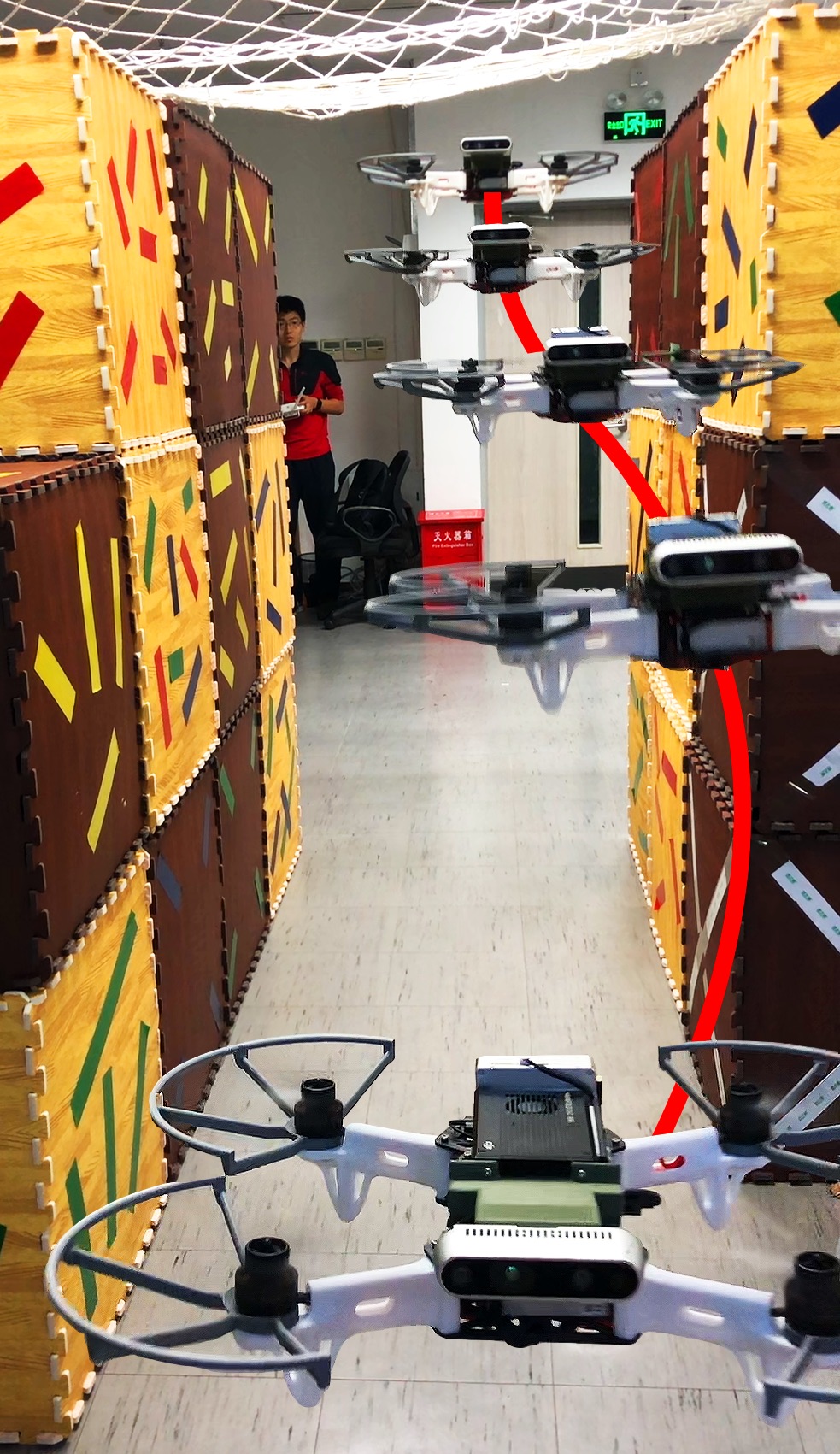}}
\subfigure[\label{fig:large_tunnel_vis} Visualization.]
{\includegraphics[height=0.43\columnwidth]{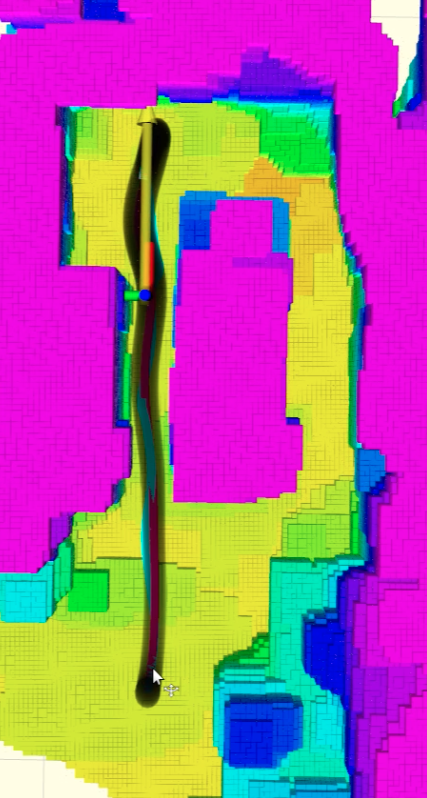}}
\subfigure[\label{fig:large_tunnel_error} The error statistics.]
{\includegraphics[height=0.33\columnwidth]{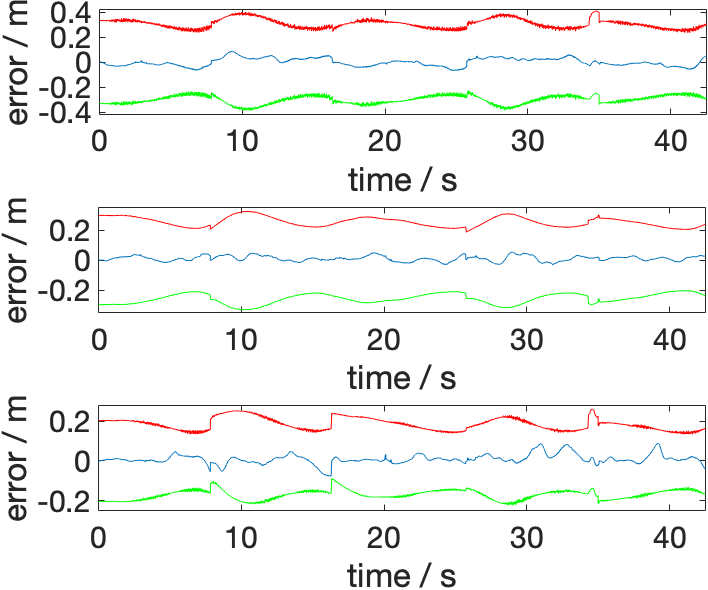}}
\end{center}
\vspace{-0.5cm}
\caption{\label{fig:large_tunnel} The composite image, visualization, and the error statistics on 3 directions for the larger quadrotor flying through a narrow gap. In the visualization, the color code indicates the height of the obstacles, the axes indicates the quadrotor's position, the yellow arrow indicates the position command and the black corridor indicates the calculated reachable sets along the trajectory according to the estimated disturbance. In the figure of the error statistics, the blue lines indicate the real error while the red and green lines indicate the estimated upper and lower bound of the error.}
\vspace{-1.0cm}
\end{figure}

\subsection{Motion Planning and Comparison}
\label{subsec:planning_result}
\begin{figure}[t]
\begin{center}         
\subfigure[\label{fig:small_composite} Snapshot.]
{\includegraphics[width=0.67\columnwidth]{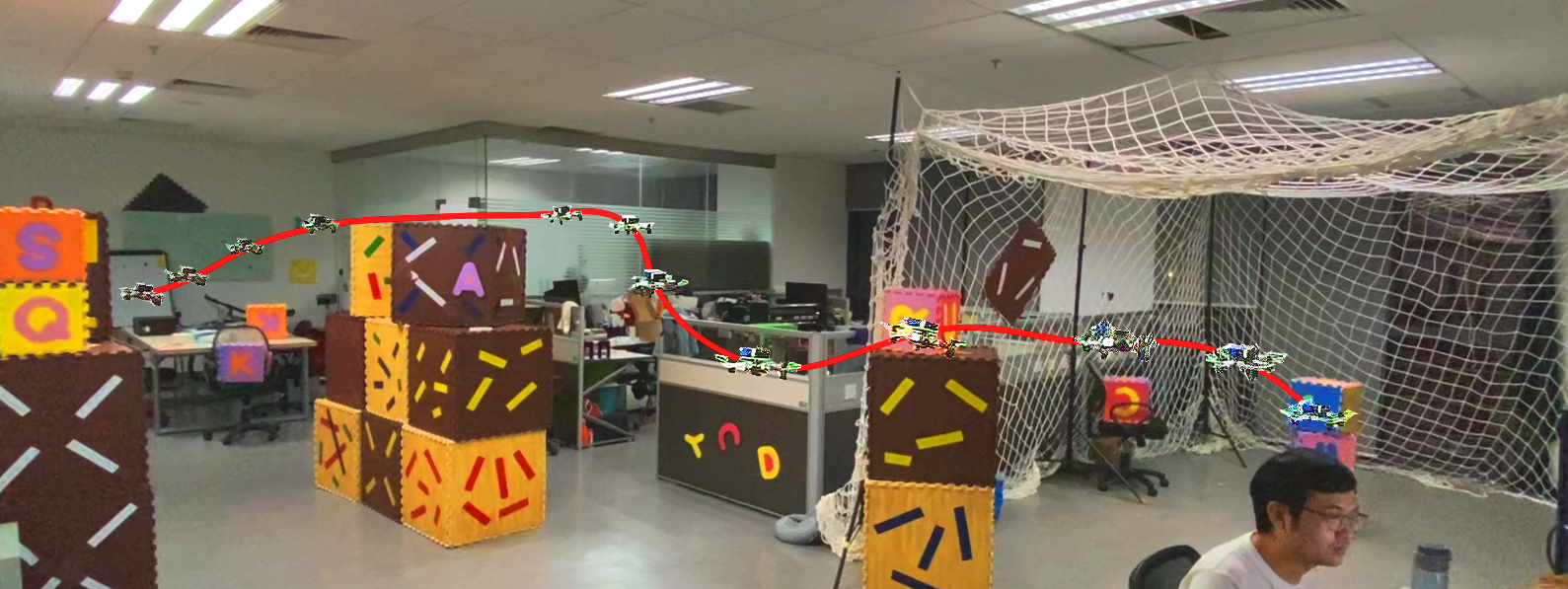}}
\subfigure[\label{fig:small_vis} Visualization.]
{\includegraphics[height=0.32\columnwidth]{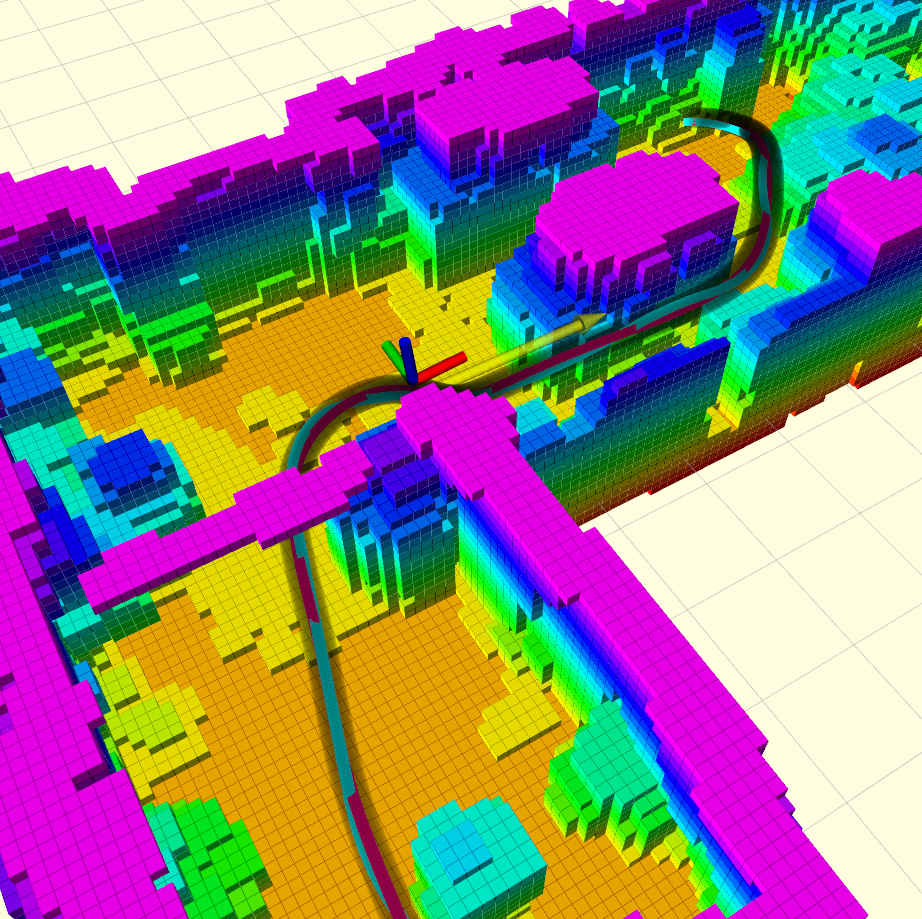}}
\subfigure[\label{fig:small_error} The error statistics.]
{\includegraphics[height=0.32\columnwidth]{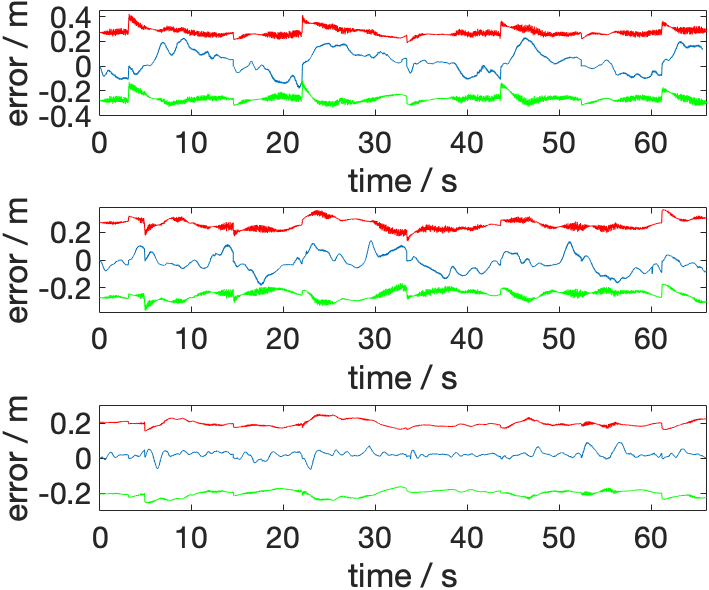}}
\vspace{-0.5cm}
\end{center}
\caption{\label{fig:small_exp} The composite image, visualization, and the error statistics on 3 directions for the smaller quadrotor flying in a complex environment. In the visualization, the color code indicates the height of the obstacles, the axes indicates the quadrotor's position, the yellow arrow indicates the position command and the black corridor indicates the calculated reachable sets along the trajectory according to the estimated disturbance. In the figure of the error statistics, the blue lines indicate the real error while the red and green lines indicate the estimated upper and lower bound of the error.}
\vspace{-0.2cm}
\end{figure}

Multiple flight tests including the narrow gap tests and the complex indoor environment tests are conducted for the verification of the proposed motion planning method. A snapshot for one of the narrow gap experiments on the larger quadrotor is shown in Fig. \ref{fig:large_composite} and the profile of the control error is shown in Fig. \ref{fig:large_tunnel_error}. Another snapshot for one of the complex environment tests on the smaller quadrotor is shown in Fig. \ref{fig:small_composite} with the control error profile shown in Fig. \ref{fig:small_error}. It can be observed that all of the position errors are within the bound of the reachable sets calculated. Furthermore, all of the errors are checked and ensured to be within the ellipsoidal reachable sets.

Moreover, comparisons with our previous motion planning method \cite{zhou2019robust} are also performed. In the experiments mentioned before, crash occurs when setting the margin $r(t)$ in Eq. \ref{eq:potential} small as a $0.2m$ constant using our previous methods. Furthermore, one of the crashes is caused by the failure of ensuring collision-free in the convex-hull of the B-spline, since our previous method enforces the collision penalty on the control points rather than the trajectory, which requires the control points to be close enough, especially when the trajectory is near the obstacles. In contrast, since the proposed method estimates the disturbance as well as ensure the clearance of the reachable sets, it does not crash or touch the obstacles during the experiments. More comparisons with our previous methods using larger margins are conducted and the results are shown in Tab. \ref{tab:benchmark_compare}. It is shown that the proposed method can obtain lower energy consumption comparing with our previous method using larger margins. The disturbance estimation generally provides a guideline for proper margin-choosing, which leads to this result.

\begin{table}[t]
\begin{center}
\caption{Comparison of Trajectory}
\vspace{-0.1cm}
\label{tab:benchmark_compare}
\begin{tabular}{|c|c|c|c|}
\hline
\multirow{2}{*}{Method} & \multirow{2}{*}{Length $(m)$} & \multirow{2}{*}{Time $(s)$} & \multirow{2}{*}{$Jerk^2$ $((m/s^3)^2)$} 
\\ &&& \\
\hline

\textbf{Proposed} & 2367.3 & 2101.5 & \textbf{5770.6} \\
\hline

$0.3m$ margin \cite{zhou2019robust} & 2335.6 & 2033.2 & 8021.3 \\
\hline

$0.4m$ margin \cite{zhou2019robust} & 2410.8 & 2119.3 & 22848 \\
\hline

\end{tabular}
\end{center}
\vspace{-1.6cm}
\end{table}

\section{Conclusion}
\label{sec:conclusion}
From the hover experiments, it is shown that the indoor ego airflow disturbance can be significant meanwhile varies during slow quadrotor flights, causing severe safety issues. Therefore, in this paper, we propose a novel estimation and adaption method for indoor ego airflow disturbance of quadrotors, meanwhile apply the result on trajectory planning. The disturbance is firstly estimated according to the discretized formulation based on voxels. Then, by adopting the HJ-reachability analysis using the estimated disturbance, the approximated control error bound along a trajectory can be calculated. With the resulting FRS information, a safe motion planning framework considering the disturbance is proposed. The method is validated by multiple quadrotors of different sizes and configurations in various indoor environments. 
In this work, the static and known environment assumption is made. In the future, we plan to extend the current framework into unknown and dynamic environments, making the quadrotor to automatically adapt the disturbance and fly safely in those scenarios.


\newlength{\bibitemsep}\setlength{\bibitemsep}{.0238\baselineskip}
\newlength{\bibparskip}\setlength{\bibparskip}{0pt}
\let\oldthebibliography\thebibliography
\renewcommand\thebibliography[1]{%
  \oldthebibliography{#1}%
  \setlength{\parskip}{\bibitemsep}%
  \setlength{\itemsep}{\bibparskip}%
}

\bibliography{ICRA2021_Luqi} 

\begin{thebibliography}{10}
\providecommand{\url}[1]{#1}
\csname url@rmstyle\endcsname
\providecommand{\newblock}{\relax}
\providecommand{\bibinfo}[2]{#2}
\providecommand\BIBentrySTDinterwordspacing{\spaceskip=0pt\relax}
\providecommand\BIBentryALTinterwordstretchfactor{4}
\providecommand\BIBentryALTinterwordspacing{\spaceskip=\fontdimen2\font plus
\BIBentryALTinterwordstretchfactor\fontdimen3\font minus
  \fontdimen4\font\relax}
\providecommand\BIBforeignlanguage[2]{{%
\expandafter\ifx\csname l@#1\endcsname\relax
\typeout{** WARNING: IEEEtran.bst: No hyphenation pattern has been}%
\typeout{** loaded for the language `#1'. Using the pattern for}%
\typeout{** the default language instead.}%
\else
\language=\csname l@#1\endcsname
\fi
#2}}

\bibitem{chong2018act}
C.~{Huang}, F.~{Gao}, J.~{Pan}, Z.~{Yang}, W.~{Qiu}, P.~{Chen}, X.~{Yang},
  S.~{Shen}, and K.~{Cheng}, ``Act: An autonomous drone cinematography system
  for action scenes,'' in \emph{Proc. of the {IEEE} Intl. Conf. on Robot. and
  Autom. ({ICRA})}, 2018, pp. 7039--7046.

\bibitem{manyam2017surveillance}
S.~G. {Manyam}, S.~{Rasmussen}, D.~W. {Casbeer}, K.~{Kalyanam}, and
  S.~{Manickam}, ``Multi-uav routing for persistent intelligence surveillance
  reconnaissance missions,'' in \emph{Proc. of the Intl. Conf. on Unma. Air.
  Syst.({ICUAS})}, 2017, pp. 573--580.

\bibitem{luqi2018collaborative}
L.~Wang, D.~Cheng, F.~Gao, F.~Cai, J.~Guo, M.~Lin, and S.~Shen, ``A
  collaborative aerial-ground robotic system for fast exploration,'' in
  \emph{Proc. of the Intl. Sym. on Exp. Robot. ({ISER})}, 2018, pp. 59--71.

\bibitem{zhou2019robust}
B.~Zhou, F.~Gao, L.~Wang, C.~Liu, and S.~Shen, ``Robust and efficient quadrotor
  trajectory generation for fast autonomous flight,'' \emph{IEEE Robotics and
  Automation Letters ({RA-L})}, vol.~4, no.~4, pp. 3529--3536, 2019.

\bibitem{fei2017iros}
F.~Gao, Y.~Lin, and S.~Shen, ``Gradient-based online safe trajectory generation
  for quadrotor flight in complex environments,'' in \emph{Proc. of the
  {IEEE/RSJ} Intl. Conf. on Intell. Robots and Syst.({IROS})}, Sept 2017, pp.
  3681--3688.

\bibitem{liu2017planning}
S.~Liu, M.~Watterson, K.~Mohta, K.~Sun, S.~Bhattacharya, C.~J. Taylor, and
  V.~Kumar, ``Planning dynamically feasible trajectories for quadrotors using
  safe flight corridors in 3-d complex environments,'' \emph{IEEE Robotics and
  Automation Letters ({RA-L})}, pp. 1688--1695, 2017.

\bibitem{lee2016robust}
S.~J. Lee, S.~Kim, K.~H. Johansson, and H.~J. Kim, ``Robust acceleration
  control of a hexarotor uav with a disturbance observer,'' in \emph{Proc. of
  the {IEEE} Control and Decision Conf. ({CDC})}.\hskip 1em plus 0.5em minus
  0.4em\relax IEEE, 2016, pp. 4166--4171.

\bibitem{hoseong2019frs}
H.~{Seo}, D.~{Lee}, C.~Y. {Son}, C.~J. {Tomlin}, and H.~J. {Kim}, ``Robust
  trajectory planning for a multirotor against disturbance based on
  hamilton-jacobi reachability analysis,'' in \emph{Proc. of the {IEEE/RSJ}
  Intl. Conf. on Intell. Robots and Syst.({IROS})}, 2019, pp. 3150--3157.

\bibitem{betz1937ground}
A.~Betz, ``The ground effect on lifting propellers,'' 1937.

\bibitem{cheeseman1955effect}
I.~Cheeseman and W.~Bennett, ``The effect of ground on a helicopter rotor in
  forward flight,'' 1955.

\bibitem{eberhart2017modeling}
G.~M. Eberhart, ``Modeling of ground effect benefits for multi-rotor small
  unmanned aerial systems at hover,'' Ph.D. dissertation, Ohio University,
  2017.

\bibitem{conyers2019empirical}
S.~A. Conyers, ``Empirical evaluation of ground, ceiling, and wall effect for
  small-scale rotorcraft,'' 2019.

\bibitem{powers2013influence}
C.~Powers, D.~Mellinger, A.~Kushleyev, B.~Kothmann, and V.~Kumar, ``Influence
  of aerodynamics and proximity effects in quadrotor flight,'' in \emph{Proc.
  of the Intl. Sym. on Exp. Robot. ({ISER})}.\hskip 1em plus 0.5em minus
  0.4em\relax Springer, 2013, pp. 289--302.

\bibitem{danjun2015autonomous}
L.~Danjun, Z.~Yan, S.~Zongying, and L.~Geng, ``Autonomous landing of quadrotor
  based on ground effect modelling,'' in \emph{2015 34th Chinese Control
  Conference (CCC)}.\hskip 1em plus 0.5em minus 0.4em\relax IEEE, 2015, pp.
  5647--5652.

\bibitem{johnson2012helicopter}
W.~Johnson, \emph{Helicopter theory}.\hskip 1em plus 0.5em minus 0.4em\relax
  Courier Corporation, 2012.

\bibitem{kushleyev2013towards}
A.~Kushleyev, D.~Mellinger, C.~Powers, and V.~Kumar, ``Towards a swarm of agile
  micro quadrotors,'' \emph{Auton. Robots}, vol.~35, no.~4, pp. 287--300, 2013.

\bibitem{robinson2014computational}
D.~C. Robinson, H.~Chung, and K.~Ryan, ``Computational investigation of micro
  rotorcraft near-wall hovering aerodynamics,'' in \emph{Proc. of the Intl.
  Conf. on Unma. Air. Syst.({ICUAS})}.\hskip 1em plus 0.5em minus 0.4em\relax
  IEEE, 2014, pp. 1055--1063.

\bibitem{girard2005reachability}
A.~Girard, ``Reachability of uncertain linear systems using zonotopes,'' in
  \emph{International Workshop on Hybrid Systems: Computation and
  Control}.\hskip 1em plus 0.5em minus 0.4em\relax Springer, 2005, pp.
  291--305.

\bibitem{kurzhanski2000ellipsoidal}
A.~B. Kurzhanski and P.~Varaiya, ``Ellipsoidal techniques for reachability
  analysis,'' in \emph{International Workshop on Hybrid Systems: Computation
  and Control}.\hskip 1em plus 0.5em minus 0.4em\relax Springer, 2000, pp.
  202--214.

\bibitem{ian2005hj}
I.~M. {Mitchell}, A.~M. {Bayen}, and C.~J. {Tomlin}, ``A time-dependent
  hamilton-jacobi formulation of reachable sets for continuous dynamic games,''
  \emph{{IEEE} Trans. Automatic Control. ({TAC})}, vol.~50, no.~7, pp.
  947--957, 2005.

\bibitem{le2012sequential}
J.~Le~Ny and G.~J. Pappas, ``Sequential composition of robust controller
  specifications,'' in \emph{Proc. of the {IEEE} Intl. Conf. on Robot. and
  Autom. ({ICRA})}.\hskip 1em plus 0.5em minus 0.4em\relax IEEE, 2012, pp.
  5190--5195.

\bibitem{herbert2017fastrack}
S.~L. Herbert, M.~Chen, S.~Han, S.~Bansal, J.~F. Fisac, and C.~J. Tomlin,
  ``Fastrack: A modular framework for fast and guaranteed safe motion
  planning,'' in \emph{Proc. of the {IEEE} Control and Decision Conf.
  ({CDC})}.\hskip 1em plus 0.5em minus 0.4em\relax IEEE, 2017, pp. 1517--1522.

\bibitem{fridovich2018planning}
D.~Fridovich-Keil, S.~L. Herbert, J.~F. Fisac, S.~Deglurkar, and C.~J. Tomlin,
  ``Planning, fast and slow: A framework for adaptive real-time safe trajectory
  planning,'' in \emph{Proc. of the {IEEE} Intl. Conf. on Robot. and Autom.
  ({ICRA})}.\hskip 1em plus 0.5em minus 0.4em\relax IEEE, 2018, pp. 387--394.

\bibitem{majumdar2017funnel}
A.~Majumdar and R.~Tedrake, ``Funnel libraries for real-time robust feedback
  motion planning,'' \emph{Intl. J. Robot. Research ({IJRR})}, vol.~36, no.~8,
  pp. 947--982, 2017.

\bibitem{dolgov2010path}
D.~Dolgov, S.~Thrun, M.~Montemerlo, and J.~Diebel, ``Path planning for
  autonomous vehicles in unknown semi-structured environments,'' \emph{Intl. J.
  Robot. Research ({IJRR})}, vol.~29, no.~5, pp. 485--501, 2010.

\bibitem{mueller2015computationally}
M.~W. Mueller, M.~Hehn, and R.~D'Andrea, ``A computationally efficient motion
  primitive for quadrocopter trajectory generation,'' \emph{{IEEE} Trans.
  Robot. ({TRO})}, vol.~31, no.~6, pp. 1294--1310, 2015.

\bibitem{quinlan1993elastic}
S.~Quinlan and O.~Khatib, ``Elastic bands: Connecting path planning and
  control,'' in \emph{Proc. of the {IEEE} Intl. Conf. on Robot. and Autom.
  ({ICRA})}.\hskip 1em plus 0.5em minus 0.4em\relax IEEE, 1993, pp. 802--807.

\bibitem{zhu2015convex}
Z.~Zhu, E.~Schmerling, and M.~Pavone, ``A convex optimization approach to
  smooth trajectories for motion planning with car-like robots,'' in
  \emph{Proc. of the {IEEE} Control and Decision Conf. ({CDC})}.\hskip 1em plus
  0.5em minus 0.4em\relax IEEE, 2015, pp. 835--842.

\bibitem{qin2019a}
T.~Qin, J.~Pan, S.~Cao, and S.~Shen, ``A general optimization-based framework
  for local odometry estimation with multiple sensors,'' 2019.

\bibitem{wang2019real}
K.~Wang, F.~Gao, and S.~Shen, ``Real-time scalable dense surfel mapping,'' in
  \emph{Proc. of the {IEEE} Intl. Conf. on Robot. and Autom. ({ICRA})}.\hskip
  1em plus 0.5em minus 0.4em\relax IEEE, 2019, pp. 6919--6925.

\end{thebibliography}
\end{document}